\documentclass[runningheads]{llncs}
\usepackage{graphicx}
\usepackage[square,sort,comma,numbers]{natbib}
\usepackage{wrapfig}
\usepackage{url}
\begin{document}
\title{
A Multi-class Approach -- Building a Visual Classifier based on Textual Descriptions using Zero-Shot Learning}
\titlerunning{Visual Classification using Zero-Shot Learning}
%
\author{Preeti Jagdish Sajjan \and Frank G. Glavin}
\authorrunning{P. J. Sajjan \and F. G. Glavin}
%
\institute{School of Computer Science, \\College of Science and Engineering, \\National University of Ireland, Galway, Ireland.\\
\email{p.sajjan1@nuigalway.ie, frank.glavin@nuigalway.ie}\\
}
\maketitle              
\begin{abstract}
Machine Learning (ML) techniques for image classification routinely require many labelled images for training the model and while testing, we ought to use images belonging to the same domain as those used for training. In this paper, we overcome the two main hurdles of ML, i.e. scarcity of data and constrained prediction of the classification model. We do this by introducing a visual classifier which uses a concept of transfer learning, namely Zero-Shot Learning (ZSL), and standard Natural Language Processing techniques. We train a classifier by mapping labelled images to their textual description instead of training it for specific classes. Transfer learning involves transferring knowledge across domains that are similar. ZSL intelligently applies the knowledge learned while training for future recognition tasks. ZSL differentiates classes as two types: seen and unseen classes. Seen classes are the classes upon which we have trained our model and unseen classes are the classes upon which we test our model. The examples from unseen classes have not been encountered in the training phase. Earlier research in this domain focused on developing a binary classifier but, in this paper, we present a multi-class classifier with a Zero-Shot Learning approach.



\end{abstract}

\section{Introduction}
In this work, we develop a visual classifier capable of classifying images based on their textual descriptions. This is made possible by mapping visual features from images and textual features from the descriptions to the semantic label space of the classes. We do this by using Zero-Shot Learning (ZSL). As the name would suggest, Zero-Shot Learning can be defined as a setup where the model is given certain classes during its testing phase which were not included in the training phase. In simpler terms, ZSL intelligently allows the model to recognise and classify objects that it has never seen before with some degree of certainty. We consider earlier work by Elhoseiny \citep{Elhoseiny} as a base reference.

\subsection{Motivation and Dataset}
Concepts of transfer learning, such as Zero-Shot Learning and Few-Shot Learning, have achieved greater visibility producing significant research in the past decade. The main motivation for applying the concepts of Zero-Shot Learning is to achieve a model that can classify the images into certain categories without being trained upon it. The key idea of ZSL is to explore and exploit the knowledge of how an unseen class is semantically related to seen classes. We present a classifier that is built using various emerging Computer Vision, Deep Learning, and NLP techniques. The novelty here is in creating a state-of-the-art Zero-Shot Learning multi-class classification model that learns to map images to their textual descriptions and, in turn, to their class labels. While our model is trained on a much smaller dataset, when compared to earlier work, we still achieve promising results. \\
\indent The dataset that we use for building our model is the \emph{Caltech Birds dataset} CUB200-2011 Birds \citep{birds-dataset} which has 11788 images of 200 species of bird. Textual descriptions for each of the above 200 classes are obtained from earlier research \citep{Elhoseiny} with information extracted manually from Wikipedia. Examples of the textual descriptions and the Caltech Birds dataset, organised with respect to the class labels, are shown below in Figure \ref{Fig1}.
\begin{figure}[ht]
    \centering
    \includegraphics[width=0.6\textwidth]{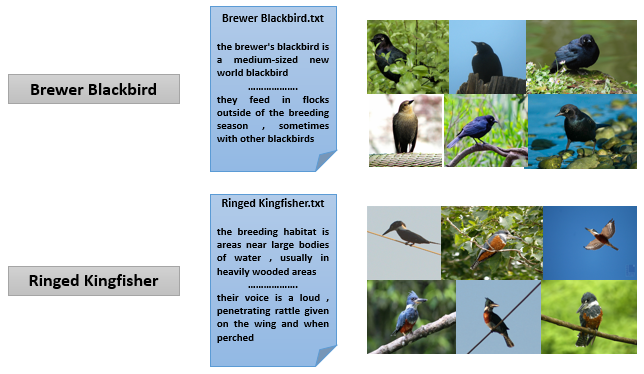}
    \caption{Samples from the dataset [Class Label, Text Description, Images]}
    \label{Fig1}
\end{figure}
\section{Background Information and Related Work}
\subsection{Transfer Learning}
Transfer Learning (TL) involves transferring the knowledge learnt in one domain to another domain. There are many real-world applications in which collecting new training data can be difficult or expensive. Transfer learning aims to reduce the need to collect such training data by transferring knowledge between the task domains. Pan and Yang \citep{Pan} presented a survey on TL and noted that its motivation was initially discussed in a \emph{NIPS-95} workshop as \emph{Learning to Learn}, which focused on the need for machine learning methods to learn, and then use this learned knowledge, when a sample from an unseen category or domain appears.
\subsection{Applying Zero-Shot Learning to Image Classification}
Lack of training data for each class and learning, both local and global, features for a group of images makes image classification using Zero-Shot Learning a challenging task. Li et al. \citep{Li}, formulate ZSL as a conditioned image classification problem where they aim to classify the visual features using a classifier that has learned from semantic descriptions. ZSL models typically learn the mapping function that maps the feature space to the semantic vector space. Since with ZSL, the model only has visibility of instances from training classes, it suffers a \emph{projection domain shift}. This problem of ZSL was first identified by Fu et al. \citep{Fu}. Kodirov et al. \citep{Kodirov} present Semantic Autoencoders, where an encoder aims to project the visual feature vectors to the semantic vector space and a decoder aims to reconstruct the original visual feature vectors from the semantic vector space. Once the model is trained, the authors retrieve the encoder model which establishes an optimal mapping function as a solution to this. Akata et al. \citep{Akata} use images from seen classes and semantic attributes, from both seen and unseen classes, to learn two dictionaries (``\emph{coupled dictionary}''), that can sparsely represent the visual and semantic feature vectors of an image. They also provide an \emph{attribute-aware} system to solve domain shift and the \emph{hubness}\footnote{A problem in which a few select words (or hubs) are too close to many others, especially in high dimensional spaces.} problems of ZSL. Akanksha et al. \citep{Akanksha} propose a Semantically Aligned Bias Reducing (SABR) approach that focuses on overcoming the \emph{hubness problem} by learning a \emph{latent space} which is responsible for preserving the semantic relation between the labels and then encoding the discriminating information of the classes.
\subsection{Mapping Textual Descriptions to Images}
Humans can write a summary of events seen in an image to provide a better understanding of that image. Wang J et al. \citep{Wang} describe a system to establish a link from an image to a sentence using a score from the comparison made between the context vector of an image and the context vector of a sentence. Later, other researchers \citep{Wang2020} 
presented a system that predicted the natural language descriptions automatically from the image input. This is made possible using recognition algorithms and exploiting statistics behind parsing huge text data. Carrara et al. \citep{Carrara} proposed \emph{Text2Vis}, a neural network that is capable of generating a visual representation in the visual feature space of ImageNet from a short textual description. This concept serves as another end of our objective for classifying images. In more recent work, Otto et al. \citep{Otto2019} outlined an approach to understand, categorise, and predict the semantic relations of an image to text. Here, the authors derive a categorisation of eight semantics in image-text pairs and illustrate how they can systematically be characterised by a set of metrics. They make use of a Deep Learning system to predict the classes using \emph{multimodal embeddings}. 
\subsection{Image Classification based on Textual Descriptions} 
Many earlier approaches, relevant to this paper, include texts from the web to train and predict a Zero-Shot Learning classifier. Elhoseiny et al. \citep{Elhoseiny} proposed an optimised formulation that combines knowledge transfer techniques and a regression function to predict a visual classifier. Ba et al. \citep{Ba} considered deep neural networks to predict convolutional classifiers by using text features to find the optimal weights for the layers of a deep neural network. This approach was reported to give a noticeable improvement in zero-shot classification. Qiao et al. \citep{Qiao} revisited the importance of regularisation on ZSL. They show that applying the attribute-based formulation to text achieves better performance. 
\section{Model Architecture and Methodology}
In this section, we describe our model architecture in detail. This is divided into three components, i) extracting features from images, ii) extracting features from text documents, and iii) building a ZSL model that learns to classify the image as a result of mapping these attributes.
\subsection{Image Feature Extraction}
Every image is stored in a machine in the form of a matrix, where each element in this matrix holds the image's pixel value. Any mathematical operations on an image are inherently performed on this matrix.\\
\indent From image feature extraction, we intend to collect representative data from the images. Features play a crucial role in the domain of computer vision and image processing. Keras provides a wide variety of Deep Learning models that are used for image classification, feature extraction, and transfer learning. 
We decided to use VGG16 \citep{VGG16} due to its promising results on learning critical features in images. VGG16 is a convolutional neural network with 16 layers. This model was proposed by Simonyan and Zisserman \citep{VGG16} and the authors report achieving a 92.7\% \emph{top-5} test accuracy.\\
\indent As an aside, we will explain what we mean by ‘Top-5 accuracy’. If we are testing our model with an image of ‘cat’ and the model predicts \{ ‘lion’, ‘tiger’, ‘cat’, ‘leopard’, ‘dog’ \} classes in ascending order of their probabilities then since the expected class is one among the predicted classes, we consider this as a true positive instance.
\subsection{Text Feature Extraction}
A challenging problem when dealing with text descriptions is to understand the context. Traditional NLP techniques can fail when context is important. The second component of our model is to extract features from the text. We have a total of 200 textual descriptions, one for each class. Most previous work uses Term Frequency Inverse Document Frequency (TF-IDF) to retrieve the features from the text. Some researchers also form attributes manually helping the model to focus on important aspects of the text. TF-IDF is inefficient for capturing semantics and overlooks the position of text in a document. On the other hand, manually defining the attributes makes the model unstable when it encounters the raw textual descriptions. As a solution, one of the biggest breakthroughs in understanding the context in the text is ELMo (Embeddings from Language Models), an NLP state-of-the-art framework introduced by AllenNLP \citep{ELMo}.\\
\indent ELMo is a deep contextualised word representation that captures the semantics and context of a word in a sentence \citep{ELMo}. Word vectors from ELMo are achieved as a result of the computations carried out on top of a two-layer bidirectional language model (biLM) where each layer has a forward and backward pass. This architecture converts a raw text string to word vectors with the help of a character-level CNN. These word vectors are provided as input to the first layer of biLM, the output of which forms intermediate word vectors which are then fed to the second layer of biLM. ELMo word vectors are the result of the weighted sum of raw word vectors and the two intermediate word vectors. Since this architecture works on a character level, and forms the vectors considering the entire sentence, a word can have different word vectors under different contexts \citep{Prateek}. TensorFlow Hub is the library enabling us to use ELMo in our work. 
\subsection{Building a Zero-Shot Learning Model}
The above retrieved visual feature vectors, and textual feature vectors are fed into our neural network as input.  We form two sets of classes: seen (training) and unseen (testing or zero-shot). We then form two datasets (seen and zero-shot) within our code that consist of two attributes (image features, text features) and one label (class) each. \\
\indent One might wonder, if we are not including samples from test classes during training, how is it possible for our model to predict those classes? This is what makes Zero-Shot Learning a challenging and interesting area. To build a model, we have to make sure that both the independent and dependent attributes are located in the same space. Hence, to achieve this, we explore the relationship between these attributes with the help of intermediate level semantic vector representation. This representation is introduced to enable sharing knowledge and to establish a mapping function between seen and unseen classes. These semantic representations could be achieved either by semantic attributes or by semantic word vectors. We will be focusing on semantic word vectors where we project each class label into semantic space and these projections are then used as prototypes for our Zero-Shot Learning classes. For simplicity let us call these semantic word vectors \emph{class vectors}. To obtain class vectors for each class in our dataset, we will be using Google's Word2Vec model.
\begin{figure}[ht]
    \centering
    \includegraphics[width=0.6\textwidth]{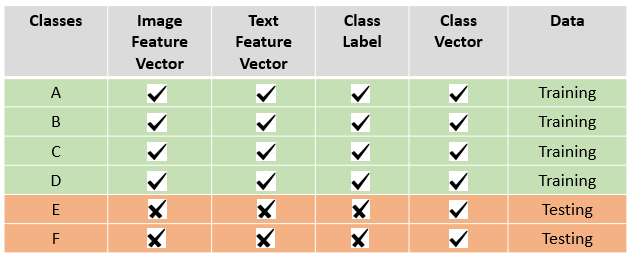}
    \caption{Visibility for training and testing our model}
\end{figure}

To summarise, for training classes, we have image features, text features, training class labels, and training class vectors. However, for testing classes, we only have their class vectors visible to our model. Note that the model has never seen any samples from testing classes and only has visibility to their class vectors.


\section{Proposed Model}
Since we aim at developing Zero-Shot Learning classifiers, we have to define a training model that learns on mapping our attributes (visual features from the image and textual features from descriptions) to their class vectors and in turn to their class labels. In order to achieve this, we use \emph{categorical cross entropy} as our loss function. If we consider visual feature vector $x_i$, corresponding textual feature vector $t_j$ and $I_{i , j}$ as the actual class label in categorical form then the loss function is computed using the following equation:
\[
Loss\ =\ - \sum_{i=1}^{C}I_{i , j}\ \log(\ \hat{y_j}\ (x_i,\ t_j))
\]
where $\hat{y_j}\ (x_i,\ t_j)$ is the $j^{th}$ scalar value in the prediction we obtain from our model.
Once we are done training our model, we pop out the last layer so that we get class vectors as predictions. When a sample from an unseen class is given to this network, we will be able to obtain a semantic word vector as an output. Since we know that if a vector is nearer to a certain class vector then the sample has a higher probability of belonging to that class. By measuring the output vector's distance to all other class vectors, we will be able to perform classification.\\
\indent Let $x^{(n1)}$ denote visual feature vector for an image, $y^{(n2)}$ denote textual feature vector for the corresponding textual description, and $l \in \{1, ..., C\}$ denotes the class labels, and $(n1, n2)$ represents vector dimensionality, $x, y \in {R}^d$, and $C$ is the total number of classes we have in our dataset. We split this data into training and testing with $M, N$ instances respectively, where $p \in l_M \not\in l_N$ or $p \in l_N \not\in l_M$. Therefore, now we have a training set $D_{train}\ =\ \{x_{train}^{(n1)},\ y_{train}^{(n2)},\ l_{train}\}$ and testing set $D_{test}\ =\ \{x_{test}^{(n1)},\ y_{test}^{(n2)},\ l_{test}\}$. The model will be trained on a training set $D_{train}$ and testing set $D_{test}$ will not be seen until the model is being tested. 
\subsection{Training Model}
We are considering 171 classes for training with 60 images per class. Once we have visual feature vectors and corresponding textual feature vectors, we feed both these vectors to our neural network. The network developed consists of 10 hidden layers, one semantic vector space layer with 300 neurons (embedding size) followed by the output layer with 171 neurons (number of training classes). All the hidden layers are equipped with \emph{‘relu’} activation and the final output layer is with \emph{‘softmax’}. While training the network, the model learns to map both the feature vectors to corresponding class labels with the help of class vectors in semantic vector space. So, in turn, it learns to map the feature vectors to their class vectors. The below illustration explains this clearly.
\begin{figure}[ht]
    \centering
    \includegraphics[width=0.8\textwidth]{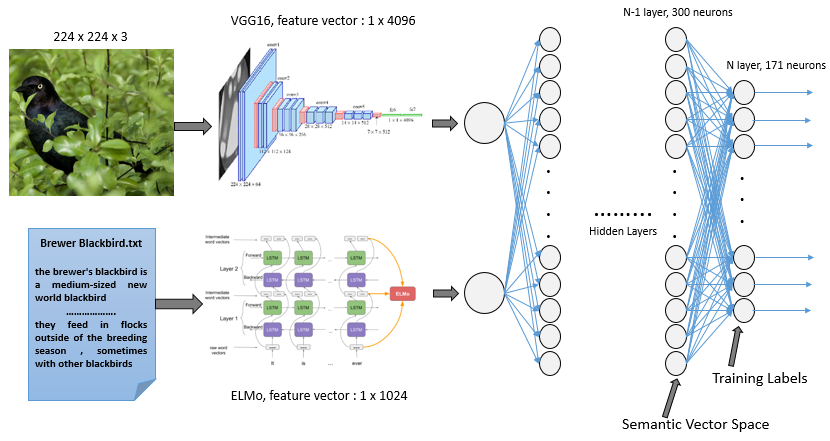}
    \caption{Proposed Training Model with image and text input}
\end{figure}
\emph{Brewer Blackbird} is one of the training classes and a model with a sample of this class is illustrated above. As we can see, the image for this class is fed to the VGG16 which yields us an image feature vector of $1 \times 4096$ dimensions. On the other hand, a corresponding textual document is fed to the ELMo which yields us a textual feature vector of $1 \times 1024$ dimensions. These two feature vectors form attributes of our model and are fed to train our neural network, the mapping function. Once we finish training our network, we save the model along with its weights which are used during testing.
\subsection{Testing Model}
We are considering 25 classes (each with 60 images) for testing our model. When we say unseen classes, samples belonging to these 25 classes have not been encountered before. The visual features and textual features of these classes are very new to the model. The process is like that of training in the beginning. To perform ZSL, we pop out the final softmax layer from the training model discussed above and let the model predict the class vector in the semantic vector space. Now, our model will predict for any pair of attributes, a class vector of $(1 \times 300\ dimensions)$ in the semantic vector space.\\
\indent \textbf{\emph{Semantic Vector Space Mapping:}} Let us here see how we retrieve the class label when our model is predicting only the class vector. We make use of K-dimensional trees, popularly known as \emph{KDTrees}. KDTrees are defined as a binary search tree where data in each node is a K-Dimensional point in space. We can derive the class labels by providing a query to this tree instance along with the number of nearest nodes or vectors we intend to consider (here k = 10, 5, 1). These class labels are arranged in increasing order of their distance from the predicted class vector.
\begin{figure}
    \centering
    \includegraphics[width=0.8\textwidth]{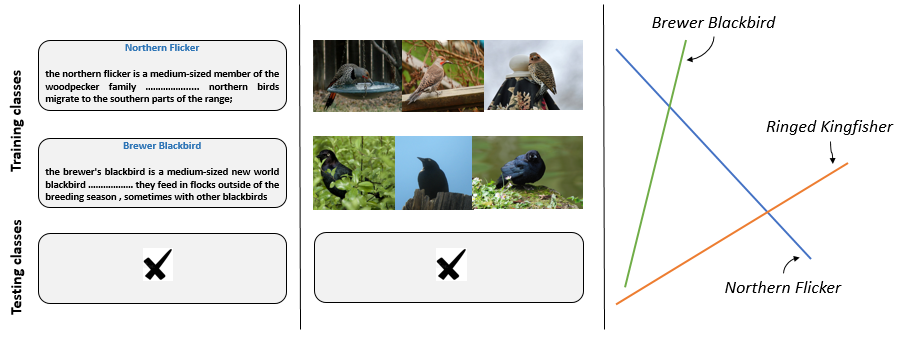}
    \caption{Semantic Vector Space illustrating the visibility of the model during training}
\end{figure}

\section{Experimental Settings}
\subsection{Image Feature Extraction}
When compared to that of the traditional hand-engineered feature extraction techniques, CNNs outperforms others by learning the complex and crucial feature representation from the raw image pixels. The image feature vectors obtained from the CNNs hold local and spatial information from the pixels of the image. Therefore, we import the pre-trained VGG16 from Keras models and by popping the final sigmoid layer we extract $1 \times 4096$ dimension feature vectors. The optimiser used while training this model is stochastic gradient descent with a learning rate of 0.1 and loss being categorical cross-entropy. The hyper-parameters are set to their default values. We load the images using the Keras \emph{load image} package and preprocess it before feeding it to the network. 
Consider we have $N$ images belong to $C$ classes where label $l\ = \{1, ...., C\}$. $\forall n \in N,\ VGG16(n)$ will obtain a feature vector $x^{(n1)}$, where $n1$ is the dimensionality of the vector i.e $1 \times 4096$. Now once we extract feature vectors for all the images, we will be having $x_N^{(n1)}$. 
\subsection{Textual Feature Extraction}
The features from the textual descriptions are obtained from ELMo using \emph{TensorFlow Hub}. We extract the embeddings in the form of a dictionary. To successfully extract features from the text, we have to clean it and perform lemmatisation (normalisation) where the text converts each word to their base form. We provide this preprocessed text as an input to the word embedding instance (ELMo) which outputs the corresponding feature vector.\\
\indent As specified, we have 200 classes in our dataset and the textual description for each of these 200 classes are purely extracted from Wikipedia. Let us assume that we have M documents where $M\ =\ C$ classes with labels $l\ =\ \{1, ...., C\}$. $\forall m \in M,\ ELMo(m)$ will obtain for us an embedding or textual feature vector $y^{(n2)}$, where $n2$ is the dimensionality of the vector i.e $1 \times 1024$. Now, once we extract feature vectors for all the textual documents, we have $y_M^{(n2)}$.
\subsection{ZSL Model} 
The above two feature vectors $x^{(n1)}, y^{(n2)}$ are the two independent variables or attributes to our model. In our experimental results, we found that reducing the dimensionality of the visual feature vector from $1 \times 4096$ to $1 \times 1024$ increased the model performance by 5\%. This dimension reduction is performed by a neural network with 3 hidden layers equipped with a \emph{‘relu’} activation function and a final \emph{‘softmax’} layer. To avoid the model overfitting, we perform batch normalisation and add dropout layers to the network. We fed this reduced visual feature vector to the neural network along with textual feature vectors. This neural network is developed with five hidden layers, followed by the semantic vector space layer and then by the final softmax layer. The hidden layers and the semantic vector space layer are equipped with a \emph{‘relu’} activation function. \\
\indent Let us now explain how we get class vectors for each class label in the semantic vector space. From our experimental results, we found that \emph{Word2Vec} establishes better embeddings when compared to that of \emph{GloVe}. This is mostly because we wish to extract embeddings for the scientific names of the species. Word2Vec makes it possible for us to have embeddings for 196 of 200 classes so we remove the classes that it cannot handle. As a result, we have 11547 images belonging to 196 classes. Word2Vec provides us an embedding of $1 \times 300$ dimensions. We obtain an instance of Word2Vec from gensim models and once we extract the class vectors from this instance, we save it locally as a numpy file. We load this numpy file later while mapping the class vectors to the class labels. The final layer of the model is responsible for this mapping function. This is achieved with the help of \emph{customised kernel initialisation} offered by Keras. Here we load the saved numpy file with class labels as keys and class vectors as values. And for every instance in the training dataset, we return the vector associated with the class label (key) which makes it possible for our model to learn a mapping of the semantic vectors to the class labels.
\section{Results and Observations}
\subsection{Model Evaluation}
The evaluation metric we are using is \textbf{top-k accuracy}. According to Ba et al. \citep{Ba}, the best way to evaluate a multi-class classifier is by sorting the final prediction score $\hat{y_c}$ obtained from the model. We deal with Zero-Shot Learning on top of multi-class classification which increases the complexity due to the much larger prediction space.
\begin{table}[ht]
\centering
\begin{tabular}{ |c|c|c|c| } 
\hline
  & Top-10 Accuracy(\%) & Top-5 Accuracy(\%) & Top-1 Accuracy(\%) \\ [0.5ex] 
 \hline \hline
 Seen classes & $99.9\pm\ 0.01$ & $99\pm\ 0.5$ & $97\pm\ 0.5$ \\ 
 \hline
 Unseen classes & $43\pm\ 0.7$ & $32\pm\ 0.5$ & $19\pm\ 0.8$ \\ 
 \hline
\end{tabular}
\caption{Proposed Zero-Shot Learning Model performance on Caltech Birds dataset}
\label{table1}
\end{table}

Table \ref{table1} illustrates the results achieved by our model for classifying the samples for both seen and unseen class categories for the Caltech Birds dataset. Seen samples are tested using the usual machine learning technique of splitting the data (here belonging to training classes) into training and testing and evaluating the performance of the model. The above results demonstrate that our approach of developing a multi-class Zero-Shot Learning classifier can classify the images belonging to classes it has seen and also to the classes it has never seen before. 
\subsection{Performance Evaluation}
Many earlier research approaches use articles from the web to develop a Zero-Shot Learning classifier. One such approach, which gave a noticeable improvement in zero-shot classification is where Ba et al. \citep{Ba} considered deep neural networks to predict convolutional classifiers. Since this is most relevant to our work, we wish to discuss the results of this approach. The authors considered 160 seen and 40 unseen classes from the available 200 classes that are randomly assigned. Ba et al. achieved the highest accuracy for the \textbf{unseen} classes, a \textbf{top-5 accuracy} of 42.8\%, and \textbf{top-1 accuracy} of 12\%. Also, the author mentions the performance of the model for \textbf{seen} classes, a \textbf{top-5 accuracy} of 66.8\%. The `training-testing' split for the seen class's evaluation is not reported in the work.\\
\indent Using the same dataset, we have developed a model using an altogether different approach. We consider a word embedding \emph{ELMo} for extracting textual features and pre-trained \emph{VGG16}, pre-trained on the ImageNet dataset, for extracting visual feature vectors. Both are then fed to the neural network as attributes. Our model considers 171 classes as seen and 25 classes as unseen. The model achieves for \textbf{unseen} classes a \textbf{top-5 accuracy} of $32\pm 0.5\%$ and \textbf{top-1 accuracy} of $19\pm 0.8$\%. To test the performance of our model for \textbf{seen} classes, we have considered \emph{70:30} random split in the training dataset (samples from 171 training classes) and, our model could achieve \textbf{top-5 accuracy} of $99\pm 0.5$\% and \textbf{top-1 accuracy} of $97 \pm 0.5$\%. Since we were able to find the optimal weights for our multi-class classifier, the proposed model performs extremely well while classifying images belonging to both seen and unseen classes . \\
\indent Top-1 accuracy is when the model assigns the highest probability for the actual class. Since we are able to achieve promising results for the most critical predictions i.e top-1 accuracy, for the unseen classes and with the fact that the model is trained on a much smaller dataset, it is evident that the model proposed in this paper is efficient and achieves promising results. 
\subsection{Model Predictions}
Below is a visualisation of the predictions produced by our model. 
\begin{figure}[ht]
    \centering
    \includegraphics[width=0.7\linewidth]{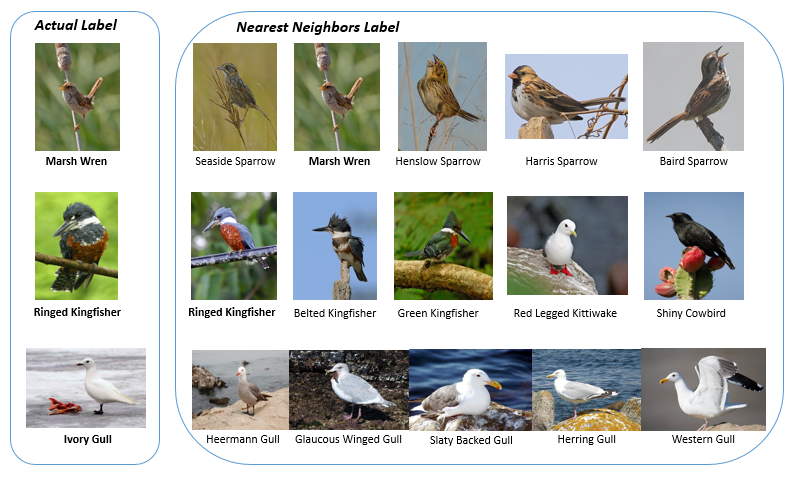}
    \caption{\label{fig:Fig9}Model predictions on testing of zero-shot classes.}
\end{figure}
This visualisation illustrates the top-5 accuracy metric, explained earlier where the leftmost class in the \emph{nearest neighbor labels} has the highest similarity score, and this similarity score decreases to the right. ‘Marsh Wren’, ‘Ringed Kingfisher’, and ‘Ivory Gull’ all belong to our testing classes i.e. the classes whose samples are never seen by the model during training. The model performance is satisfactory for the first two examples. We can see that the third example is one such instance where our model is unable to predict the exact class label but the classes which are chosen by the model under top-5 nearest neighbors belong to the same parent class (‘Gull’). 
\section{Conclusion}
In this paper, we introduce a multi-class Zero-Shot Learning model that learns to predict the label for images belonging to unseen classes from their Wikipedia articles. We have developed a deep neural network to establish a mapping function that maps the textual feature vectors from the raw Wikipedia articles and visual feature vectors from images to a semantic space with the help of semantic vectors derived from the class labels. This can also be interpreted as the ability of the model to intelligently apply the knowledge acquired over training classes with the help of an objective function. \\
\indent We demonstrated that our model outperformed an existing zero-shot model on the top-1 accuracy metric on the CUBird dataset using only raw images and Wikipedia articles.

\bibliographystyle{splncs04}
\bibliography{References}
\end{document}